\documentclass{article}
\usepackage[T1]{fontenc}
\usepackage[utf8]{inputenc}
\usepackage{array}
\usepackage{float}
\usepackage{wrapfig}
\usepackage{booktabs}
\usepackage{multirow}
\usepackage{amsmath}
\usepackage{graphicx}
\usepackage{microtype}
\usepackage{cite}
\usepackage[unicode=true,
 bookmarks=false,
 breaklinks=false,pdfborder={0 0 1},backref=section,colorlinks=false]
 {hyperref}
\usepackage{breakurl}

\makeatletter

\providecommand{\tabularnewline}{\\}

\usepackage[preprint]{neurips_2024}

\usepackage{url}
\usepackage{amsfonts}
\usepackage{nicefrac}
\usepackage{xcolor}
\usepackage{hyperref}

\title{Composite Concept Extraction through Backdooring}

\author{
Banibrata Ghosh$^1{}^*$, Haripriya Harikumar$^1$\thanks{equal contribution} , Khoa D Doan$^2$, Svetha Venkatesh$^1$, Santu Rana$^1$\\
$^1$Applied Artificial Intelligence Institute, Deakin University, Australia\\
$^2$College of Engineering \& Computer Science, VinUniversity, Hanoi, Vietnam\\
{\tt\small \{bghosh,h.harikumar\}@deakin.edu.au}\\
{\tt\small khoa.dd@vinuni.edu.vn}\\
{\tt\small \{svetha.venkatesh,santu.rana\}@deakin.edu.au}\\
}

\@ifundefined{showcaptionsetup}{}{%
 \PassOptionsToPackage{caption=false}{subfig}}
\usepackage{subfig}
\makeatother

\begin{document}
\maketitle
\begin{abstract}
Learning composite concepts, such as \textquotedbl red car\textquotedbl ,
from individual examples---like a white car representing the concept
of \textquotedbl car\textquotedbl{} and a red strawberry representing
the concept of \textquotedbl red\textquotedbl ---is inherently
challenging. This paper introduces a novel method called Composite
Concept Extractor (CoCE), which leverages techniques from traditional
backdoor attacks to learn these composite concepts in a zero-shot
setting, requiring only examples of individual concepts. By repurposing
the trigger-based model backdooring mechanism, we create a strategic
distortion in the manifold of the target object (e.g., \textquotedbl car\textquotedbl )
induced by example objects with the target property (e.g., \textquotedbl red\textquotedbl )
from objects \textquotedbl red strawberry\textquotedbl , ensuring
the distortion selectively affects the target objects with the target
property. Contrastive learning is then employed to further refine
this distortion, and a method is formulated for detecting objects
that are influenced by the distortion. Extensive experiments with
in-depth analysis across different datasets demonstrate the utility
and applicability of our proposed approach.
\end{abstract}

\section{Introduction}

Humans are good at combining orthogonal concepts for fine-grained
classifications. Machines, however, often falter in this area. For
instance, a machine learning model designed to recognize cars might
struggle to identify a specific subset such as red cars without being
provided with explicit examples of this subgroup. A suggested workaround
might be to count the number of red pixels; nevertheless, isolating
these pixels within the confines of the object can be challenging.
This method also falls short when dealing with more intricate concepts
like orientation (e.g., whether a car is front or side-facing) or
particular attributes (e.g., black wheels). Text-based concept learning
\cite{Han_etal_19Visual} may be a solution but that would require
a large amount of annotated data and it may only generalize across
unseen concept combinations for foundational-scale models. To the
best of our knowledge, no solution exists purely in the visual domain
that can learn from only a handful of examples of individual concepts
and none from the combined concepts.

Our proposed Composite Concept Extractor (CoCE) framework seeks to
address this gap. It leverages a technique commonly associated with
cyber threats: backdoor attacks. Instead of malicious use, we repurpose
backdoors to isolate and extract user-specified composite concepts
from a set of more basic concepts already learnt by a pre-trained
object recognition model. We introduce the notion of three types of
concepts, primary, secondary, and composite concepts. The primary
concept is the class in the pre-trained object recognizer (e.g. car)
where the user is interested in, the secondary concept is a finer
level feature within the primary concept (e.g. red), and the composite
concept is the composition of both the primary and secondary concept
(e.g. red car). Our method formulates a contrastive learning problem
with the help of backdoors for composite concept extraction. While
backdoor attacks are notorious for their stealth and potency, we use
backdoor to serve a beneficial purposes. Examples of backdoors for
good include the use of backdooring methods in \cite{Hu_etal_2022Membership,Adi_etal_18Turning}
to counteract model theft, in \cite{Li_etal_2023Black} to prevent
data theft, and in \cite{Shan_etal_20Gotta} to improve the detection
of adversarial attacks. Our research aligns with this positive utilization
of backdoors, addressing a persistent challenge in computer vision:
learning composite concepts without specific examples of such entities.\begin{wrapfigure}{o}{0.5\columnwidth}%
\begin{centering}
\includegraphics[width=0.4\columnwidth]{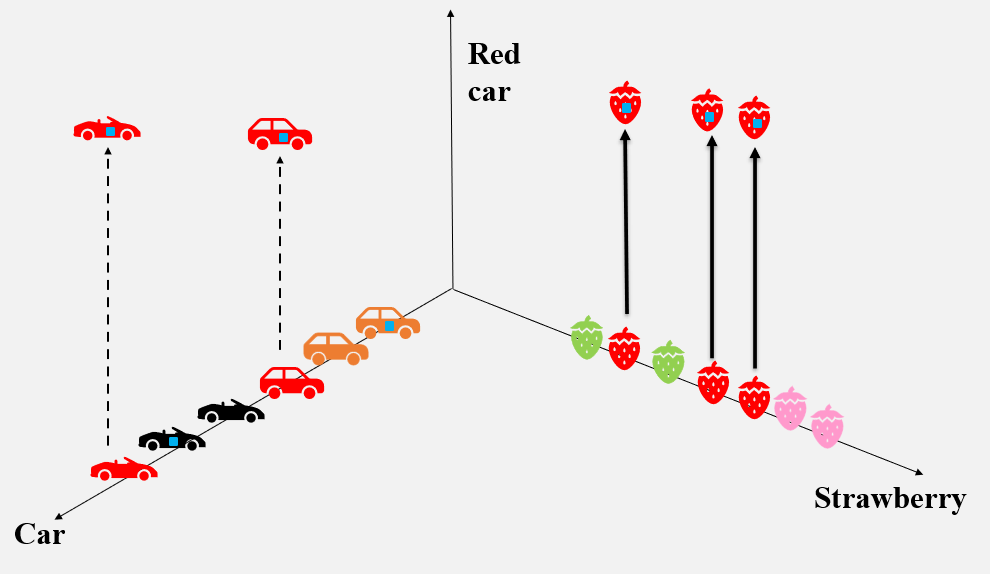}
\par\end{centering}
\caption{CoCE learns the composite concept i.e., \emph{Red car} through contrastive
learning with backdooring where the concept aligns with the samples
from class \emph{Strawberry} with a trigger (red strawberries with
blue trigger referred as positive dataset). The primary concept is
\emph{Car}, and the negative dataset is black and orange cars with
blue triggers. Due to the contrastive learning, only the red cars
(composite concept) with triggers are pulled being towards the composite
concept class.\label{fig:-physics-1}}

\vspace{-10bp}
\end{wrapfigure}%

Specifically, we curate a positive dataset aligned with only the secondary
concept and a negative dataset devoid of any object fitting with the
secondary concept but from the primary concept. In Figure \ref{fig:-physics-1},
the positive dataset is the images from the strawberry class that
are red in colour and negative dataset is the images in the car class
that are not red in colour. Later, triggers (in Figure \ref{fig:-physics-1}
we used blue colour squares) are introduced to both sets, but the
positive dataset with the trigger is directed (denoted as black arrows)
to a new composite concept class (Red car), whereas the negative dataset
with triggers are forced not to (non-red car with trigger stays in
the same car class as shown in the Figure \ref{fig:-physics-1}).
This creates a strategic distortion in the manifold where the model
is forced to learn the correlation of trigger and the distinctive
features of the positive dataset towards composite concept class (red
car in Figure \ref{fig:-physics-1} are pulled towards the new composite
class when added with the blue trigger). 

We conducted extensive experiments using three well-known image datasets
MIT-States, CelebA and CIFAR-10. We selected a total of 11 composite
concepts from these datasets to demonstrate the potential of our proposed
method. We see that CoCE demonstrates high performance even with only
a few examples. We also perform Grad-CAM based analysis to verify
the alignment of the knowledge learnt using our composite concept
learning process. Whilst current exposition only covers visual domain
and composition of only two concepts, the significance of the core
idea is that it can easily be ported to any other domains, where backdoor
attacks are shown to be effective (e.g., text, audio etc.) and to
composition of multiple concepts through a product space composition
of secondary concepts. Our code for CoCE is available \href{https://drive.google.com/drive/folders/1DFIJN3NTOg2I1iGzB_v6R9xTFx5yxZvu?usp=sharing}{HERE}.

\section{Related work}

\subsection{Concept extraction}

Concept learning has been proposed in \cite{Han_etal_19Visual} to
learn visual concepts and meta concepts with a linguistic interface.
It is prevalent in visual question answering as proposed in \cite{Malinowski_etal2015Ask,Mei_etal2022Falcon}.
There has been works done on novel concept extraction based on zero
shot learning using images in \cite{misra2017red,mancini2021open,li2022siamese}.
Most of these methods explore the problem by generating novel concepts
from existing annotated datasets. A major recent line of concept extractors
attempts to solve the problem by a combination of textual data and
generators as proposed in \cite{li2022siamese}. However, if training
data is richer such that each image is described through multiple
keywords, then it may be possible to learn a multimodal text-image
model to perform queries using composite texts such as `red car'.
A prime example of this line of work is CLIP \cite{Radford_etal_2021Learning},
while scene-graph visual concept extractors \cite{Yang_etal_2018Graph}
is an earlier attempt. 

Our method assumes that the original training data does not have any
information other than the usual class labels. Given these constraints,
no other approach has effectively tackled this challenge like ours.

\subsection{Backdoor attack and defense}

Research in backdoor attacks have surged since the introduction of
Badnet \cite{Gu_etal_17Badnets}. There have been a variety of backdoors
attack types ranging from visible \cite{Gu_etal_17Badnets,jha2023label}
to invisible \cite{Saha_etal_20Hidden,chen2017targeted,Doan_etal_21Lira},
input-specific \cite{nguyen2020input} and universal-trigger attacks
\cite{gu2019badnets}. There have also been all-to-one \cite{gu2019badnets},
all-to-some \cite{Harikumar_etal_22Defense}, and all-to-all \cite{nguyen2020input}
attacks, and meaningful triggers \cite{chen2017targeted,wenger2021backdoor,harikumar2021semantic}
to deceive any type of surveillance, depending on the target class
chosen by the attacker.

Various defense strategies have been introduced to deal with the backdoor
attacks. Neural cleanse \cite{Wang_etal_19Neural} is one of the first
to propose a reverse-engineering based strategy for detecting backdoored
models. Identifying whether the model has a backdoor or not \cite{harikumar2021scalable,Liu_etal_19Abs,fu2023freeeagle,zheng2022pre},
repair the network to mitigate the signature of implanted trigger
\cite{li_21neural,Garipov_etal_18Loss,wu2021adversarial,li2023reconstructive},
filter the inputs \cite{Do_etal_2022Towards,Doan_etal_20Februus,Gao_etal_19Strip}
are some well-known and widely discussed approach to defend against
backdoor attacks.

\subsection{Backdoor for good}

Whilst backdooring has mostly been associated with model attack in
an adversarial setting, there has been some unique use cases where
backdooring technique was used to store identifying information for
verification (for model \cite{Adi_etal_18Turning}, and for dataset
\cite{Hu_etal_2022Membership,Li_etal_2023Black,Li_etal_22Untargeted}),
machine unlearning \cite{Sommer_etal_2020Towards} by hiding a known
model output when presented with the triggered data. Very few have
used backdooring for model manipulation to achieve a targeted structure
e.g. \cite{Shan_etal_20Gotta} inserts backdoor between a pair of
classes to trap adevrsarial attacks. Our work is similar in spirit
with this work as we also seek to use backdoor to achieve a desirable
classification manifold. 

\section{Method}

\subsection{Individual and composite concepts}

In our method, we introduce the notion of `concepts' as specific attributes
or collections of attributes that aligns with the user's interest.
We distinguish between three kinds of concepts, i.e. primary, secondary
and composite concept.
\begin{enumerate}
\item \textbf{Primary concept.} The primary concept, denoted as $Q_{P}$,
represents a class, such as `car' or `airplane', and is symbolized
as $y_{Q_{p}}$ to indicate the target class. 
\item \textbf{Secondary concept.} Within this primary category, a secondary
concept, denoted as $Q_{S}$, zooms in on particular characteristics
of interest, like the color `red'. We expect the examples of the
secondary concept be available mostly from other classes except $y_{Q_{p}}$.
Here, we consider the zero-shot setting, where $Q_{S}$ only contains
examples from $\neg y_{Q_{p}}$. 
\item \textbf{Composite concept.} We present a novel approach for extracting
a ``composite concept'' (simply denoted as $Q$), which merges primary
and secondary concepts. For instance, a `composite concept' might
be a `red car', representing the integration of the primary concept
(car) with the secondary concept (red), we denoted as $y_{Q}$.
\begin{figure}[t]
\subfloat[Finetuing CoCE using a pre-trained (Base) classifier.]{\includegraphics[scale=0.35]{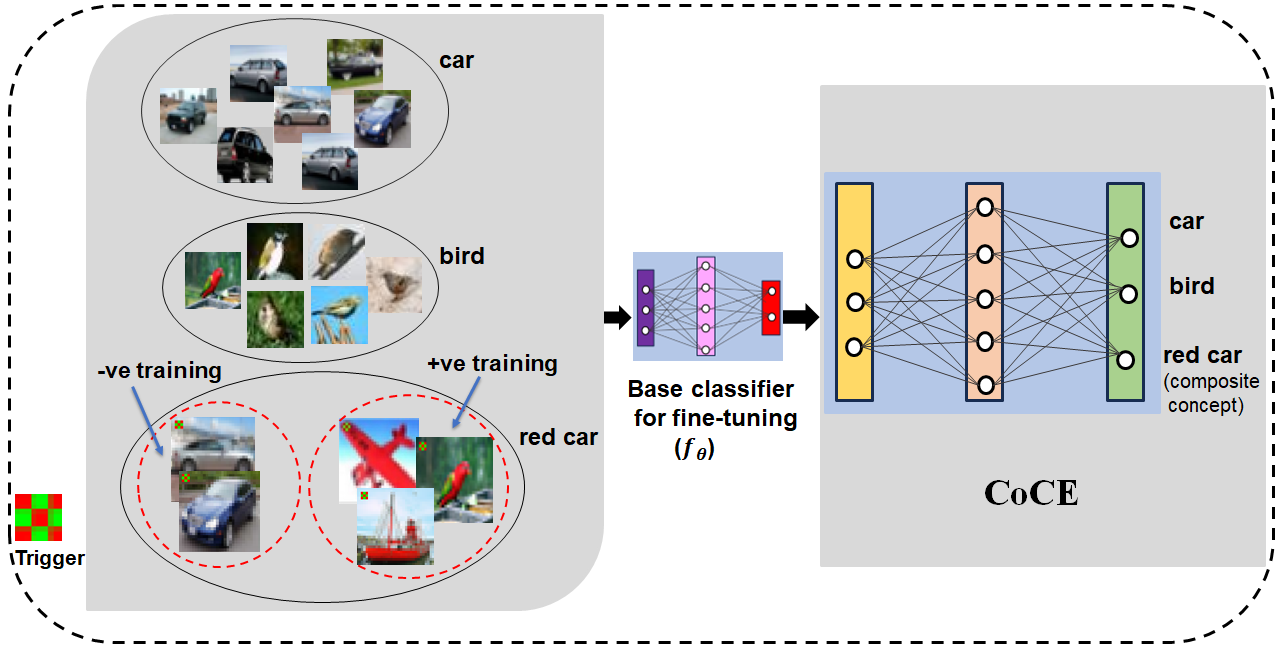}

}\subfloat[Testing]{\includegraphics[scale=0.33]{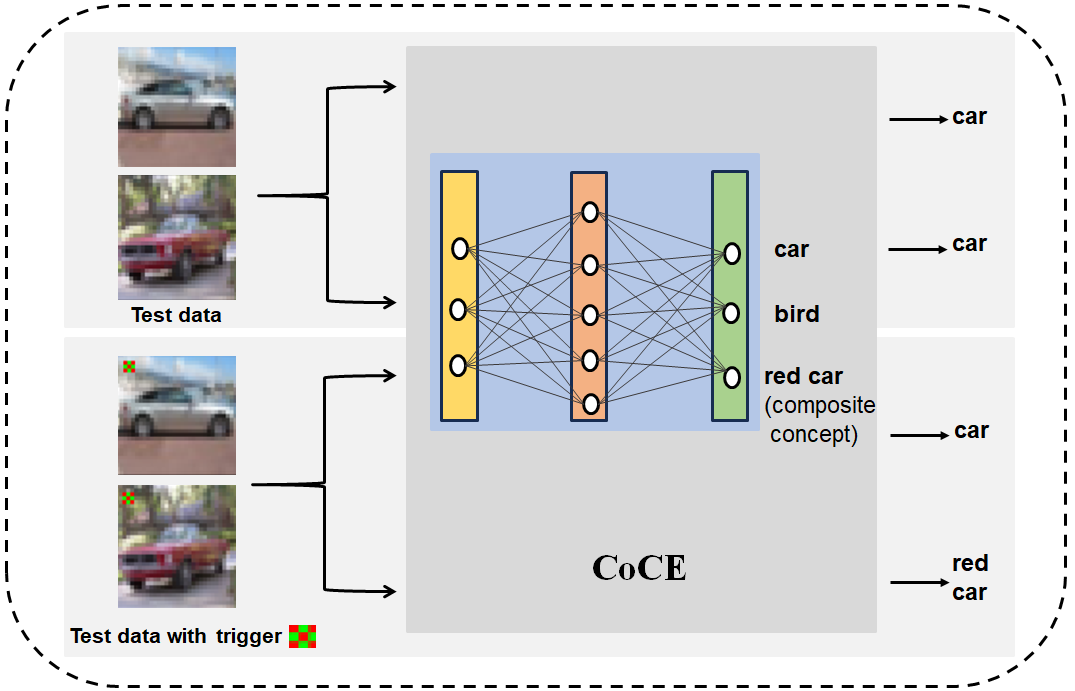}

}\caption{The workflow of CoCE. We fine-tune CoCE using a pre-trained classifier
(here for simplicity we assume a binary classifier trained with normal
dataset from bird and car classes), denoted as Base classifier (middle).
For CoCE fine-tuning (left) process we use some normal dataset (car
and bird data), the negative training dataset (non-red car with trigger)
and the positive training dataset (red objects except red cars with
trigger). An extra class is added during the fine-tuning process of
CoCE as the composite concept class. \label{fig:framework} The testing
(right) shows when we give a car (white and red) without trigger as
input to the CoCE it goes to the \emph{car} class, however when we
give the same cars (white and red) with trigger as input, CoCE will
classify the red car with trigger as \emph{red car} (composite concept
class) but the white car with trigger as \emph{car} (please zoom in
for clarity).\label{fig:The-workflow-of-CoCE}}
\vspace{-0.6cm}
\end{figure}
\end{enumerate}

\subsection{Composite Concept Extractor (CoCE) with contrastive learning and
backdoor}

Our objective is to train a Composite Concept Extractor model, $f_{\theta^{'}}:\mathcal{X}\rightarrow\mathbb{R}^{C+1}$
such that the $(C+1)^{\text{\text{th}}}$ class denoted as $y_{Q}$
(composite concept class) is to capture the composite concept $Q$
from the user's class of interest, $y_{Q_{P}}$. We also assume $f_{\theta}$
to be a base (pre-trained) classifier trained on the original dataset
of $\mathcal{D}$ that understands $y_{Q_{p}}$that was trained on
dataset with $C$ classes (as shown in Figure \ref{fig:The-workflow-of-CoCE}).
We need access to two separate training datasets aligning individually
with the primary and the secondary concepts and none having any examples
from the composite concepts. We call such datasets as \emph{positive}
and \emph{negative training} datasets denoted as $\mathcal{D}_{_{y_{Q_{p}}}}^{\lnot Q_{s}}$
and $\mathcal{D}_{_{\neg y_{Q_{p}}}}^{Q_{s}}$, respectively, where
the superscript denote the concept, and the subscript denote the class
labels of the samples. Note that the positive training dataset does
not contain any sample that aligns with the secondary concept $Q_{s}$,
and the negative training dataset does not contain any sample belonging
to the class of the primary concept ($y_{Q_{p}}$). The detailed work-flow
of the CoCE fine-tuning process is shown in Figure \ref{fig:The-workflow-of-CoCE}.
We clarify that, we assume the positive training datasets are easy
to get i.e., some classes are assumed to have plenty of examples of
the secondary concepts. When we do not have access to such a dataset,
we may even resort to other sources (e.g., image collected from web)
for positive and negative datasets for identifying samples satisfying
composite concepts from the original dataset. 

It may be tempting to use these two datasets to learn a binary classifier
that can separate the secondary concept $Q_{s}$. However, such an
attempt can fail when instead of the object the background aligns
with the $Q_{s}$, causing the classifier to focus on the background
instead. Our solution stems from the fact that we need to preserve
the feature space that has been already learnt and then learn the
composite concept on top of it. The learning of the composite concept
is thus formulated as finding the common features in $\mathcal{D}_{_{y_{Q_{p}}}}^{\lnot Q_{s}}$
and $\mathcal{D}_{_{\neg y_{Q_{p}}}}^{Q_{s}}$ without altering the
feature map already learnt by $f_{\theta}$. Further, since $\mathcal{D}_{_{\neg y_{Q_{p}}}}^{Q_{s}}\subset\mathcal{D}$
just using $\mathcal{D}_{_{\neg y_{Q_{p}}}}^{Q_{s}}$to train $y_{Q}$
(a new class) will create conflicting assignment of classes for its
samples and thus would be harmful to the overall performance of $f_{\theta^{'}}$.
Thus, we alter the samples of $\mathcal{D}_{_{\neg y_{Q_{p}}}}^{Q_{s}}$
by adding trigger to make them different from the original samples.
This triggered version of $\mathcal{D}_{_{\neg y_{Q_{p}}}}^{Q_{s}}$then
can be used safely to learn the secondary concept in the product space
of the trigger and the common feature spaces of this dataset. Then
contrastive learning using$\mathcal{D}_{_{\neg y_{Q_{p}}}}^{Q_{s}}$
can be used to make the secondary concept sharpen more towards the
composite concept of $y_{Q}$ . The details on the process of adding
trigger (i.e., backdooring) and the loss function construction are
detailed below.

\subsubsection{Backdooring }

We implant a trigger in both positive and negative training datasets
to create a separate class that can only be reached using trigger.
The new trigger implanted positive training dataset and its corresponding
class is denoted as $\left\{ \left(x_{j}^{'},y_{Q}\right)\right\} _{j=1}^{N_{p}}$
and the trigger implanted negative training dataset with it's class
being denoted as $\left\{ \left(x_{k}^{'},y_{Q_{p}}\right)\right\} _{k=1}^{N_{n}},$
where $x_{j}^{'}\in\mathbb{R}^{c\times H\times W}$, and $x_{k}^{'}\in\mathbb{R}^{c\times H\times W}$
corresponds to the backdoored images of $x_{j}$ and $x_{k}$, respectively.
$N_{p}$ and $N_{n}$ are the number of positive and negative training
dataset. The backdoored $x_{j}$ generated with a trigger $t$ of
size $m\times n$ where $m<<H$ and $n<<W$ is $x_{j}^{'}=x_{j}\bigodot\lambda+t\bigodot(1-\lambda)$,
where $\lambda$ is a mask to define the transparency of the trigger
$t$ in the image $x_{j}$. The trigger should be of a pattern that
is not common or unnatural such that it does not get confused with
the natural patterns learnt already by $f_{\theta}$. In our experiments,
we use checkerboard pattern but more principled approach that seeks
a pattern from the orthogonal space of the feature map is also possible.
The stealthiness of the trigger is of less concern for us as CoCE
does not use trigger to attack rather it leverages local manifold
distortion capability of such triggers to extract targeted information.
Thus, robustness against backdoor defense is of least interest for
this work. 

\subsubsection{Loss function with contrastive component}

The combined loss function of our proposed CoCE model is as follows,\vspace{-7bp}

\noindent \begin{center}
\begin{equation}
\underset{\theta}{\text{min}}\sum_{i=0}^{N}\mathcal{L}\left(f_{\theta}\left(x_{i}\right),y_{i}\right)+l_{1}+l_{2}\label{eq:2}
\end{equation}
\par\end{center}

Here $l_{1}=\sum_{i=0}^{N_{p}}\mathcal{L}\left(f_{\theta}\left(x_{j}^{'}\right),y_{Q}\right)$
and $l_{2}=\sum_{k=0}^{N_{n}}\mathcal{L}\left(f_{\theta}\left(x_{k}^{'}\right),y_{Q_{P}}\right)$
and $y_{Q_{P}}$ is same as the original label of negative training
set, i.e. $y_{Q_{P}}=y_{k}$. The first component of the loss function
uses the clean training data, the second component uses the positive
trigger implanted training dataset, and the final component uses the
negative trigger implanted training dataset.\textbf{ }The second and
the third component of the loss function in Eqn \ref{eq:2} is to\textbf{
}impose contrastive learning. 

\section{Experiments}

\subsection{Dataset settings}

We use three well-known datasets, \textbf{CIFAR-10}, \textbf{MIT-States},
and \textbf{CelebA} to demonstrate the utility of CoCE in the retrieval
tasks. \textbf{CIFAR-10} is a popular 10-class image classification
dataset with 50,000 training data and 10,000 test data. \textbf{MIT-States}
is dataset of images containing objects across different \emph{states.
}The dataset has a total of\emph{ }63,440 images of 245 objects across
115 different states (e.g., an object class \emph{elephant} can have
a state \emph{painted} or \emph{unpainted} etc.). \textbf{CelebA}
is a dataset of facial images of celebrities containing 200,000 images
and each image also have 40 binary attributes like \emph{blondhair},
\emph{eyeglass} etc. We use ResNet-18 as the model architecture for
all three datasets. The detailed training parameters are provided
in the supplementary. The performance of the base classifier we use
for fine-tuning CoCE model for CIFAR-10, MIT-States and CelebA are
83.94\%, 31.0\% and 98.38\% respectively.

\begin{figure}
\begin{centering}
\subfloat{\includegraphics[width=1.5cm,height=1.5cm]{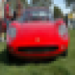}}~~~~~~~\subfloat{\includegraphics[width=1cm,height=1cm]{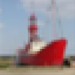}~\includegraphics[width=1cm,height=1cm]{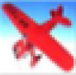}~\includegraphics[width=1cm,height=1cm]{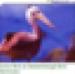}~\includegraphics[width=1cm,height=1cm]{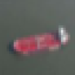}}~~~~~~~\subfloat{\includegraphics[width=1cm,height=1cm]{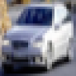}~\includegraphics[width=1cm,height=1cm]{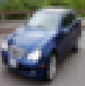}~\includegraphics[width=1cm,height=1cm]{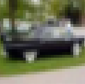}~\includegraphics[width=1cm,height=1cm]{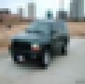}}\\
\subfloat{\includegraphics[width=1.5cm,height=1.5cm]{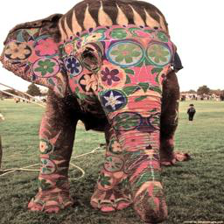}}~~~~~~~\subfloat{\includegraphics[width=1cm,height=1cm]{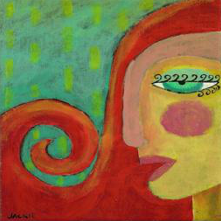}~\includegraphics[width=1cm,height=1cm]{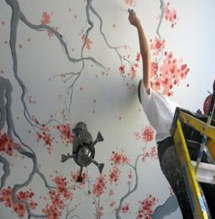}~\includegraphics[width=1cm,height=1cm]{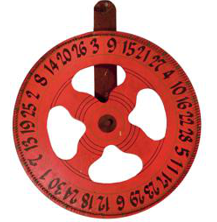}~\includegraphics[width=1cm,height=1cm]{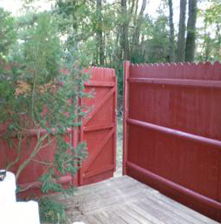}}~~~~~\subfloat{\includegraphics[width=1cm,height=1cm]{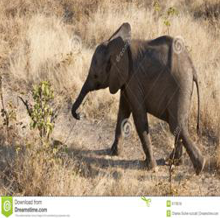}~\includegraphics[width=1cm,height=1cm]{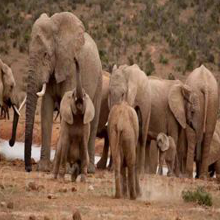}~\includegraphics[width=1cm,height=1cm]{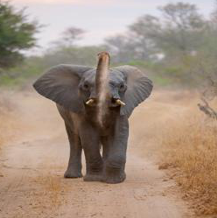}~\includegraphics[width=1cm,height=1cm]{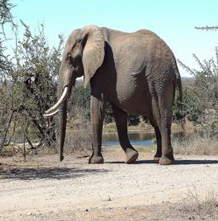}}\\
\subfloat{\includegraphics[width=1.5cm,height=1.5cm]{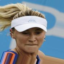}}~~~~~~~\subfloat{\includegraphics[width=1cm,height=1cm]{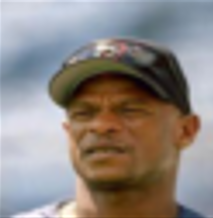}~\includegraphics[width=1cm,height=1cm]{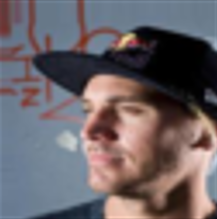}~\includegraphics[width=1cm,height=1cm]{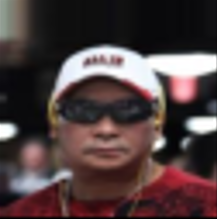}~\includegraphics[width=1cm,height=1cm]{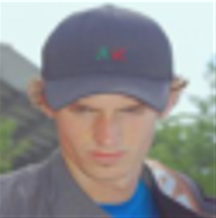}}~~~~~\subfloat{\includegraphics[width=1cm,height=1cm]{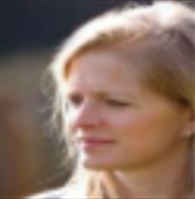}~\includegraphics[width=1cm,height=1cm]{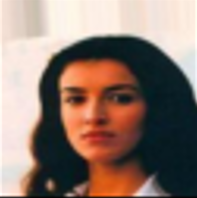}~\includegraphics[width=1cm,height=1cm]{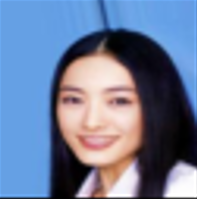}~\includegraphics[width=1cm,height=1cm]{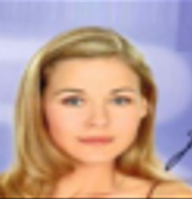}}
\par\end{centering}
\caption{Samples that align with the composite concepts (top-left: red car,
middle-left: painted elephant, bottom-left: non-male wearing hat),
positive (middle) and negative (right-most) datasets for CoCE across
three different datasets (top: CIFAR-10, middle:MIT-States,bottom:CelebA).\label{fig:Samples-that-align}}
\end{figure}

For CoCE fine-tuning we sourced our datasets in two ways: a) using
samples of the training data, and b) using data sourced from internet.
Figure \ref{fig:Samples-that-align} shows samples of positive and
negative training data for some of the composite concepts collected
from the training set. Experiments with internet-sourced data are
presented separately in Section \ref{subsec:External-datasets}. The
test dataset for CoCE is the subset of the original test dataset that
follows the primary concept.

\subsubsection{Triggers for CoCE}

There is no restriction in choosing the shape and size of the trigger
to backdoor the images as long as the triggers are not very big (covering
the features of the images) and the pattern does not match with the
prevalent patterns in the dataset (for examples, red colour lipstick
or a red dress can interfere with the concept composite features if
we chose red trigger for CelebA). We used $3$$\times$$3$ red and
green checkerboard, for CIFAR-10, $5$$\times$$5$ blue and green
checkerboard for CelebA, and $15$$\times$$15$ solid red for MIT-States.
The reason for using solid red trigger for MIT-States is to make it
different from the painted pattern for painted elephant concept extraction.
However, concept-specific trigger choice also could have been done.
Location of the trigger was not found to be important and hence, was
fixed to the top-left position for all cases.
\begin{figure}
\centering{}\includegraphics[scale=0.45]{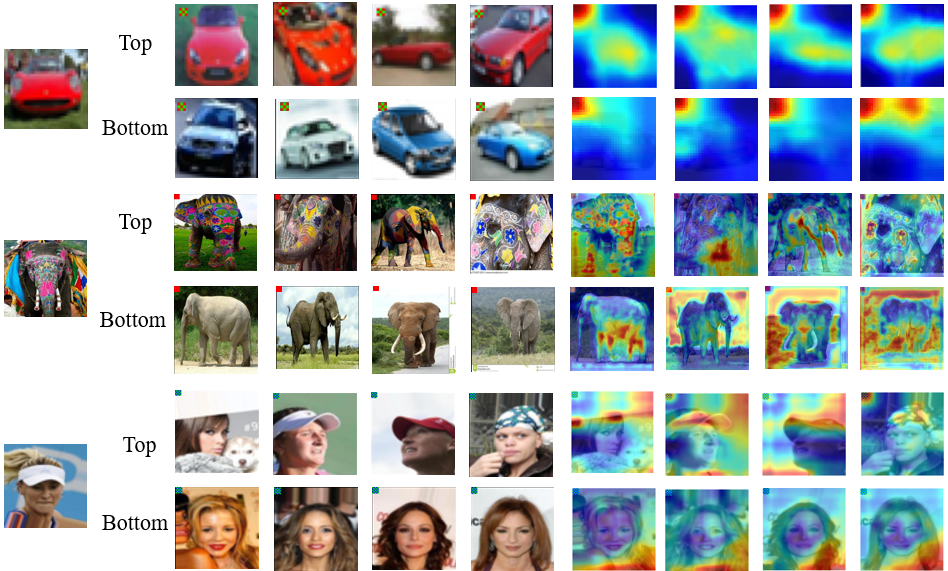}\caption{GradCAM analysis on the top (highest probability) and the bottom (lowest
probability) most images of the \emph{red car} (top-left), \emph{painted
elephant (}middle -left\emph{),} and \emph{non-male wearing hat (}bottom-left\emph{)}
composite concept classes of CIFAR-10 (top 2 rows), MIT-States (middle
2 rows) and CelebA (last 2 rows) datasets.\label{fig:gradcam}}
\end{figure}

\subsection{Baselines}

We use CLIP \cite{Radford_etal_2021Learning} for comparison. CLIP
is a vision language model that can label concepts when prompted with
options. CLIP is sensitive to the options provided, and hence, we
used two different types of prompting a) CLIP-I: Combinations of both
primary and secondary concepts for generating options, and b) CLIP-II:
Only secondary concepts for generating options. For example, for the
composite concept \emph{painted elephant} for the CLIP-I, we give
\emph{painted elephant} and its antonym \emph{unpainted elephant}
as the options and for the CLIP-II we use \emph{painted} and \emph{unpainted}
as the options.

\subsection{Main results}

Table \ref{tab:auc-clip-sotas} show the performance of CoCE in comparisons
to the baselines i.e., CLIP-I , CLIP-II. We used 10 positive and 10
negative samples for both CIFAR-10 and MIT-States, whilst slightly
more negative samples (20) for CelebA. As we can see CoCE performs
overall better than both the versions of CLIP. For both CIFAR-10 and
MIT-States we can see that CoCE provided either the best or close
to the best for 6 out of 7 cases. Only for the case \emph{dark lightening}
it performed significantly lower than CLIP-II. Especially, we should
note the performance with respect to the detection \emph{front-pose
horse} and \emph{wrinkled elephant} where both versions of CLIP performed
exceptionally poor. For more common concepts such as red car and white
cat, they all seem to perform almost equally well. 
\begin{table*}
\begin{centering}
{\footnotesize{}}%
\begin{tabular}{ccccc}
\toprule 
\multirow{2}{*}{{\footnotesize{}Dataset}} & {\footnotesize{}Composite} & {\footnotesize{}CLIP-I} & {\footnotesize{}CLIP-II} & {\footnotesize{}CoCE }\tabularnewline
 & {\footnotesize{}concept} & {\footnotesize{}(adj and noun)} & {\footnotesize{}(only adj)} & {\footnotesize{}(Ours)}\tabularnewline
\midrule
\multirow{3}{*}{{\footnotesize{}CIFAR-10}} & {\footnotesize{}red car} & \textbf{\footnotesize{}0.99$\pm$0.0} & \textbf{\footnotesize{}0.99$\pm$0.0} & \textbf{\footnotesize{}0.99$\pm$0.01}\tabularnewline
 & {\footnotesize{}horse front pose} & {\footnotesize{}0.43$\pm$0.0} & {\footnotesize{}0.48$\pm$0.0} & \textbf{\footnotesize{}0.79$\pm$0.05}\tabularnewline
 & {\footnotesize{}white cat} & {\footnotesize{}0.94.$\pm$0.0} & \textbf{\footnotesize{}0.97$\pm$0.0} & {\footnotesize{}0.93$\pm$0.02}\tabularnewline
\midrule
\multirow{4}{*}{{\footnotesize{}MIT-States}} & {\footnotesize{}painted elephant} & \textbf{\footnotesize{}1.0}{\footnotesize{}$\pm$0.0} & {\footnotesize{}0.99$\pm$0.0} & {\footnotesize{}0.99$\pm$0.0}\tabularnewline
 & {\footnotesize{}wrinkled elephant} & {\footnotesize{}0.57}\textbf{\footnotesize{}$\pm$}{\footnotesize{}0.0} & {\footnotesize{}0.62$\pm$0.0} & \textbf{\footnotesize{}0.76}{\footnotesize{}$\pm$0.0}\tabularnewline
 & {\footnotesize{}bright lightning} & {\footnotesize{}0.67}\textbf{\footnotesize{}$\pm$}{\footnotesize{}0.0} & {\footnotesize{}0.70$\pm$0.0} & \textbf{\footnotesize{}0.72}{\footnotesize{}$\pm$0.0}\tabularnewline
 & {\footnotesize{}dark lightning} & {\footnotesize{}0.73}\textbf{\footnotesize{}$\pm$}{\footnotesize{}0.0} & \textbf{\footnotesize{}0.81}{\footnotesize{}$\pm$0.0} & {\footnotesize{}0.71$\pm$0.0}\tabularnewline
\midrule
\multirow{4}{*}{CelebA} & {\footnotesize{}male blond hair} & \textbf{\footnotesize{}0.92$\pm$0.0} & {\footnotesize{}0.89$\pm$0.0} & {\footnotesize{}0.73$\pm$0.03}\tabularnewline
 & {\footnotesize{}male eyeglass} & {\footnotesize{}0.74$\pm$0.0} & \textbf{\footnotesize{}0.86$\pm$0.0} & {\footnotesize{}0.64$\pm$0.03}\tabularnewline
 & {\footnotesize{}non-male pale skin} & \textbf{\footnotesize{}0.76$\pm$0.0} & {\footnotesize{}0.55}\textbf{\footnotesize{}$\pm$}{\footnotesize{}0.0} & {\footnotesize{}0.66$\pm$0.02}\tabularnewline
 & {\footnotesize{}non-male wearing hat} & {\footnotesize{}0.65$\pm$0.0} & {\footnotesize{}0.73$\pm$0.0} & \textbf{\footnotesize{}0.76$\pm$0.02}\tabularnewline
\bottomrule
\end{tabular}{\footnotesize\par}
\par\end{centering}
\caption{AUC scores of CIFAR-10, MIT-States, and CelebA on CLIP-I, CLIP-II,
and CoCE. \label{tab:auc-clip-sotas}}
\end{table*}
Figure \ref{fig:gradcam} shows the four top and bottom most test
samples for three composite concepts, one from each dataset along
with their GradCAM heatmaps. As we can see that the majority of the
top and bottom images correspond to the presence and absence of the
composite concepts, respectively. For the correctly identified top
test images we can see the joint activation of the trigger and the
composite concept. Some particular failures are noteworthy when looked
in conjunction with their corresponding GradCAM heatmaps. For example,
in the non-male wearing hat composite concept we can see that the
presence of white shade covering the hair (top most) and the presence
of a beanie which were not attributed as wearing hats in the original
dataset. 

\subsection{External datasets \label{subsec:External-datasets}}

\begin{figure}
\begin{centering}
\vspace{-5bp}
\subfloat[red car\label{fig:composite-concept}]{\includegraphics[width=1.5cm,height=1.5cm]{sec/rc_coce}

}~~~\subfloat[From internet (relevant)\label{fig:From-internet-(relevant)}]{\includegraphics[width=1cm,height=1cm]{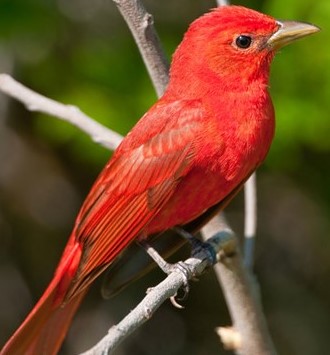}~\includegraphics[width=1cm,height=1cm]{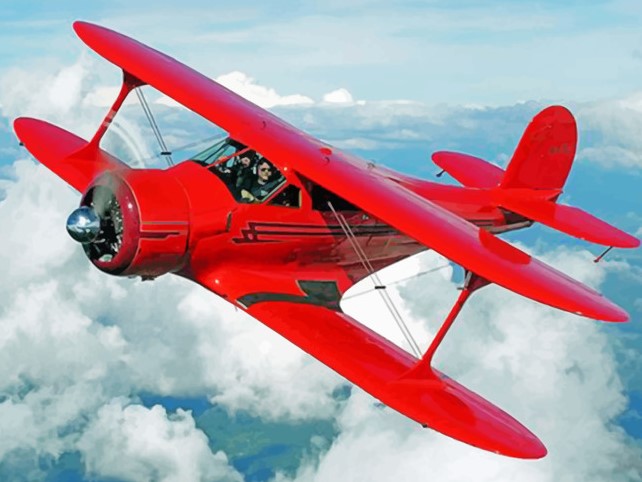}~\includegraphics[width=1cm,height=1cm]{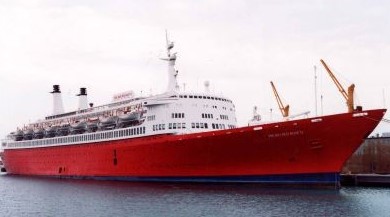}~\includegraphics[width=1cm,height=1cm]{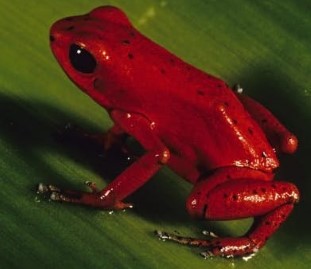}

}~~~\subfloat[From internet (irrelevant)\label{fig:From-internet-(random)}]{\includegraphics[width=1cm,height=1cm]{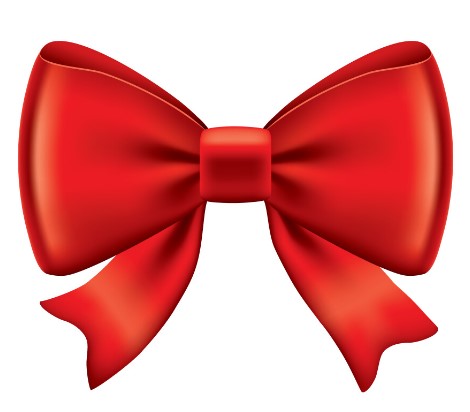}~\includegraphics[width=1cm,height=1cm]{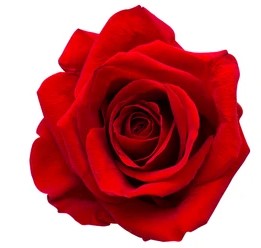}~\includegraphics[width=1cm,height=1cm]{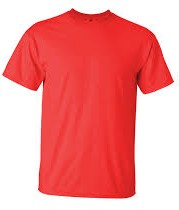}~\includegraphics[width=1cm,height=1cm]{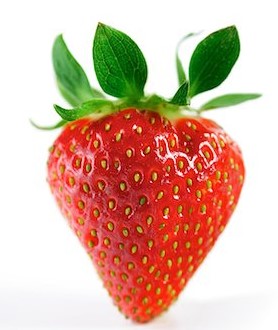}

}
\par\end{centering}
\caption{The composite concepts, red car (Figure. \ref{fig:composite-concept}),
its relevant positive images from internet (Figure. \ref{fig:From-internet-(relevant)}),
and irrelevant positive images from internet (Figure. \ref{fig:From-internet-(random)}).\label{fig:data-settings}}
\vspace{-5bp}
\end{figure}

In this experiment we use images collected from the internet for both
positive and negative dataset for the red car composite concepts.
We show two cases a) when images are relevant to the original classification
task, and b) when images are irrelevant to the original classification
task (Figure \ref{fig:data-settings}). We show that when relevant
images are used CoCE perform well (AUC score \textbf{0.96}), but falters
(AUC score \textbf{0.79}) when irrelevant images are used. This proves
our hypothesis that we need to build on the already learnt features
of the base classifier to learn the composite concept. Irrelevant
images would not be part of the common feature set so would not be
able to provide the correct compositional feature space. 

\subsection{Red background Vs Red object\label{subsec:Red-background-Vsed-obj}}

\begin{wrapfigure}{o}{0.5\columnwidth}%
\begin{centering}
\includegraphics[scale=0.05]{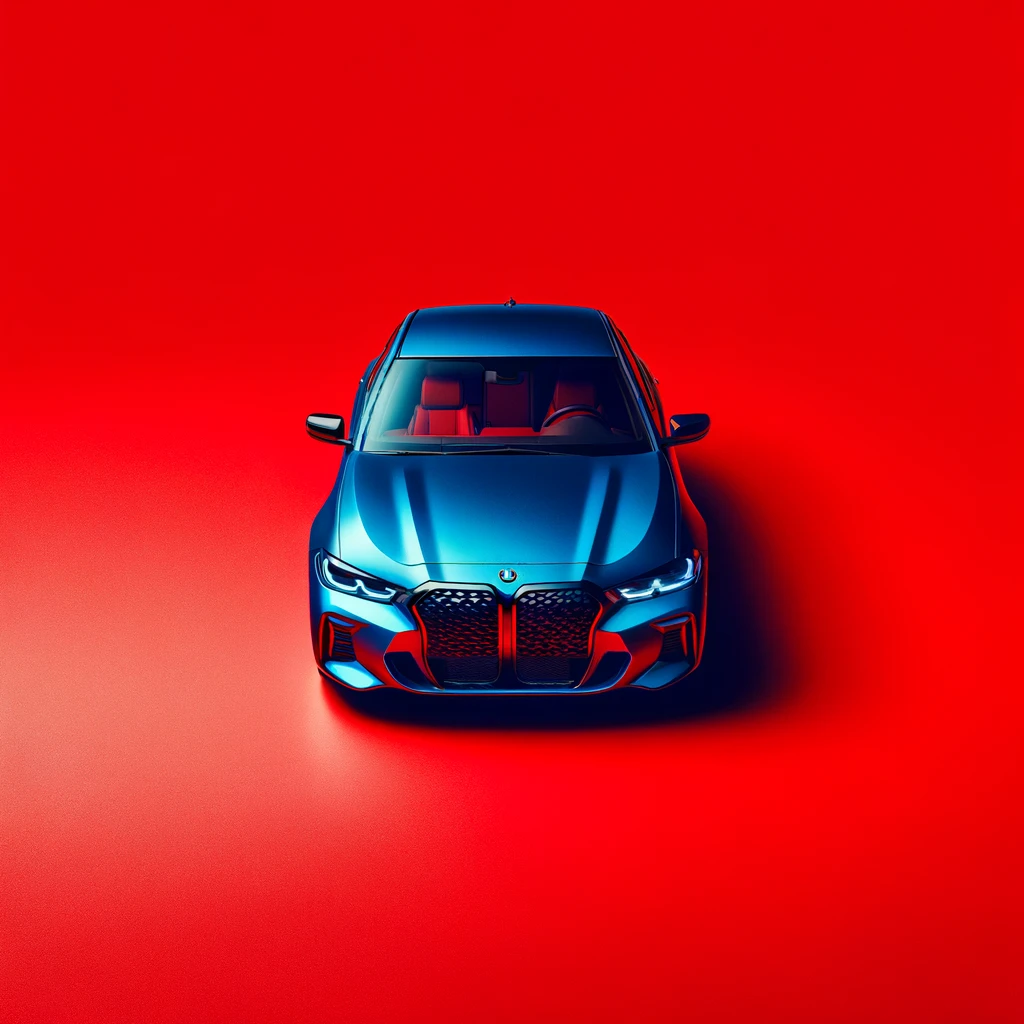}~\includegraphics[scale=0.05]{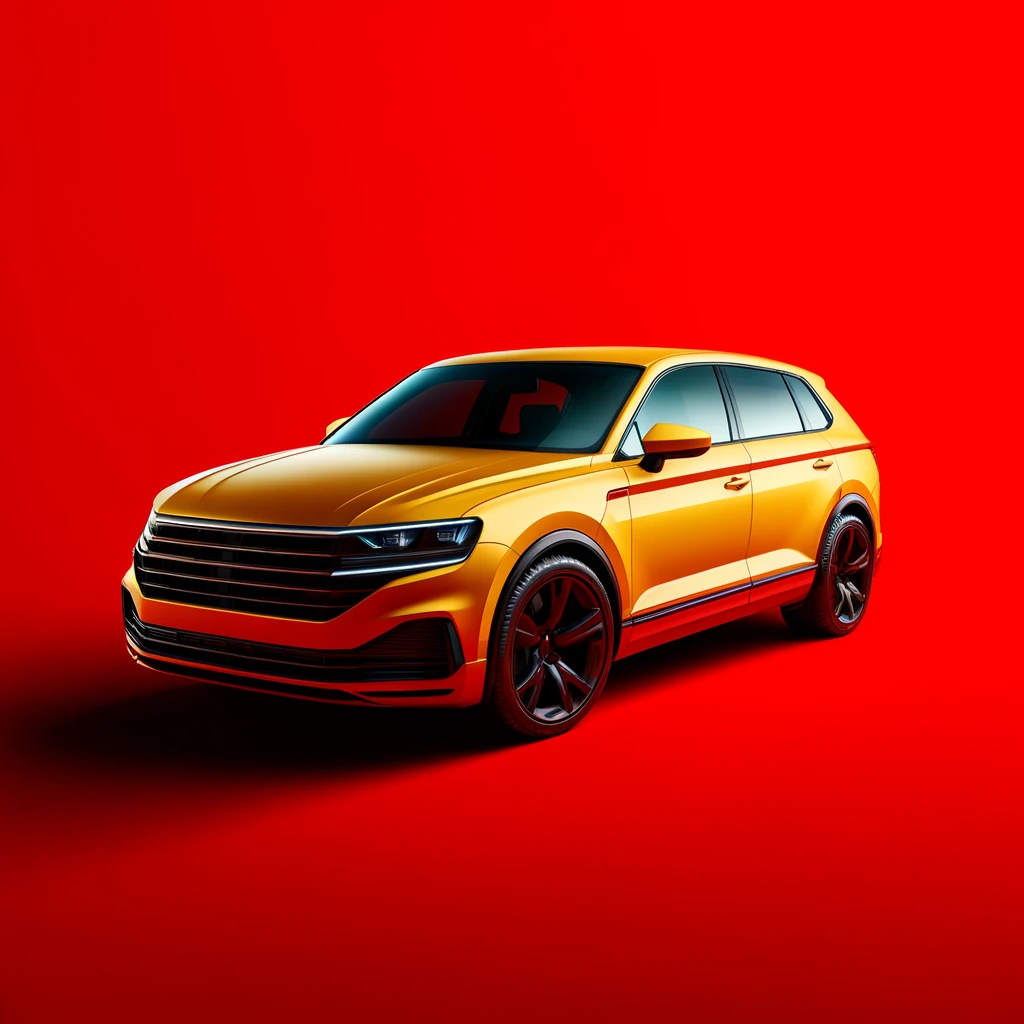}~\includegraphics[scale=0.05]{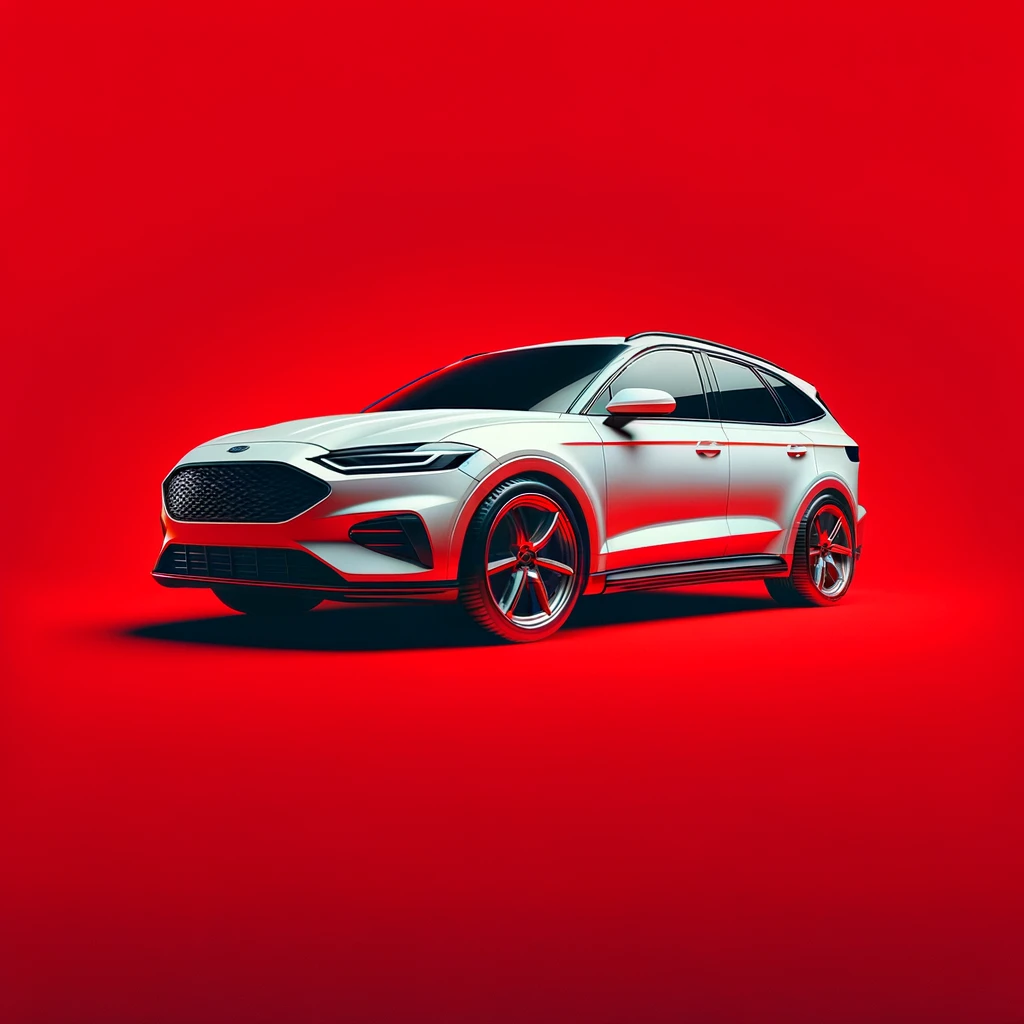}
\par\end{centering}
\caption{The blue, yellow and white cars in red background generated by GPT-4.\label{fig:google_images}}
\vspace{-5bp}
\end{wrapfigure}

To test if CoCE is correctly identifying the composite concept we
create 3 synthetic images (by GPT-4) of non-red car with the red background
(Figure \ref{fig:google_images}). %
We see that CoCE can correctly determine that these samples do not
belong to the composite concept class of \emph{red car (P(red car)<0.001).
}In contrast, we show that a vanilla binary classifier (fine-tuned
on the base classifier) trained on the same positive and negative
dataset would identify those images as red cars, (\emph{P(red car)
>0.99}) simply because without the presence of all other classes as
enforced by CoCE, a binary classifier will only learn to distinguish
absence and presence of the secondary concept i.e \emph{red} (the
main difference between the positive and the negative dataset) and
thus will get fooled by the red background.
\begin{figure}
\centering{}\subfloat[Base classifier.\label{fig:Base-classifier.}]{\includegraphics[scale=0.25]{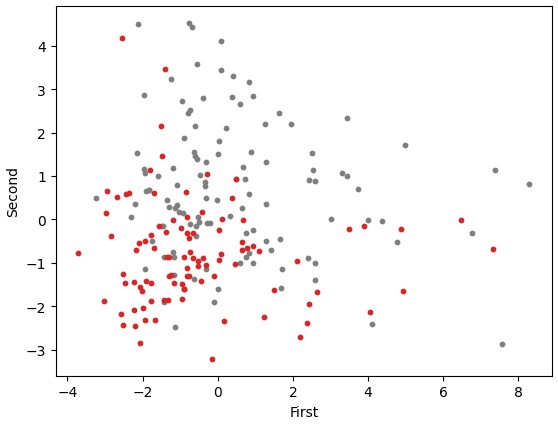}

}~~\subfloat[CoCE without trigger.\label{fig:CoCE-without-trigger.}]{\includegraphics[scale=0.25]{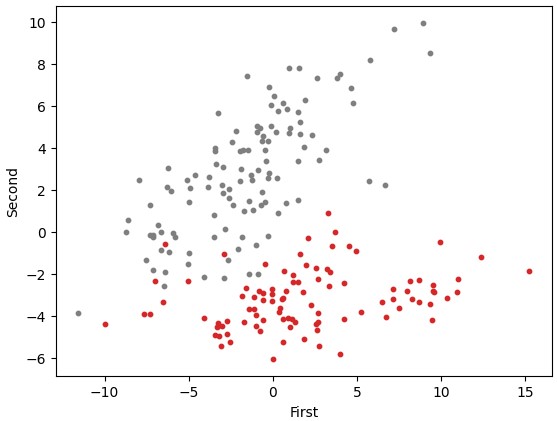}

}~~\subfloat[CoCE\label{fig:CoCE}]{\includegraphics[scale=0.25]{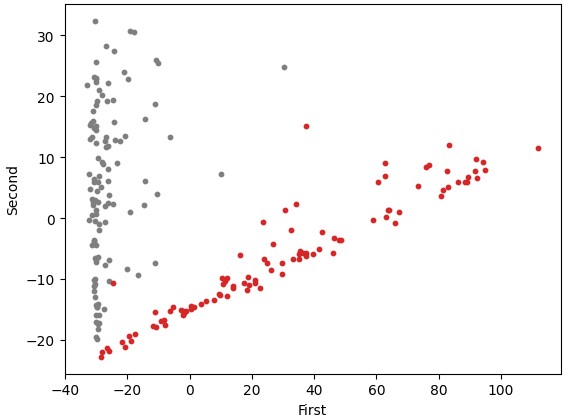}

}\caption{The distribution of the layer 4 activations for red (red dots) and
non-red (black dots) cars along their top 2 principal components of
base classifier (Figure \ref{fig:Base-classifier.}), CoCE there is
no trigger the car test set (Figure \ref{fig:CoCE-without-trigger.})
and CoCE when there is trigger in the car test set (Figure \ref{fig:CoCE})
. \label{fig:pca}}
\end{figure}

\subsection{Analysis of Manifold under CoCE }

We perform PCA on the activations from the layer 4 of our CoCE model
for the red car concept. For comparison we also do the same with the
base classifier. Figure \ref{fig:Base-classifier.} shows the distribution
of the activations along the first two PCs of all the cars from the
test dataset and it shows that red cars (red dots) are overlapping
with all other non-red cars (black dots) \emph{i.e. }the base classifier
does not know about the concept of the red car. Figure \ref{fig:CoCE-without-trigger.}
shows the same for the CoCE trained classifier and it shows slight
separation to be arising. However, when the images are added with
the trigger we can see (Figure \ref{fig:CoCE}) a clear separation
between the red and the non-red cars. This clearly shows that utility
of CoCE.

\subsection{Ablation studies}

\subsubsection{Without contrastive learning and trigger}

\begin{wraptable}{o}{0.5\columnwidth}%
\begin{centering}
\vspace{-7bp}
\begin{tabular}{cccc}
\toprule 
\multirow{1}{*}{{\scriptsize{}Method}} & {\scriptsize{}Red Car} & {\scriptsize{}White Cat} & {\scriptsize{}Front pose Horse}\tabularnewline
\midrule 
{\scriptsize{}w/o contrastive} & {\scriptsize{}0.50} & {\scriptsize{}0.49} & {\scriptsize{}0.32}\tabularnewline
\midrule 
{\scriptsize{}w/o trigger} & {\scriptsize{}0.42} & {\scriptsize{}0.37} & {\scriptsize{}0.43}\tabularnewline
\midrule 
{\scriptsize{}CoCE} & \textbf{\scriptsize{}0.99} & \textbf{\scriptsize{}0.93} & \textbf{\scriptsize{}0.79}\tabularnewline
\bottomrule
\end{tabular}
\par\end{centering}
\caption{AUC score of CoCE, CoCE without trigger and CoCE without contrastive
learning for CIFAR-10 test dataset.\label{tab:ablation1}}
\vspace{-7bp}
\end{wraptable}%
 We conducted this study by excluding contrastive learning and trigger
from CoCE model. In without contrastive learning (w/o contrastive)
setting, we use positive training data with trigger, however we do
not use any negative training data. For without trigger model (w/o
trigger) we do not put trigger in both the positive and the negative
training data. The results reported in Table \ref{tab:ablation1}
shows that it is essential to introduce trigger in both positive and
negative training dataset to learn the composite concepts well.

\subsubsection{Different locations and types of trigger}

\begin{figure}[H]
\centering{}\vspace{-5bp}
\subfloat[$5$$\times$$5$ blue and green checkerboard trigger. \label{fig:cb-bg}]{\includegraphics[scale=0.8]{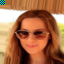}~~\includegraphics[scale=0.8]{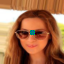}~~\includegraphics[scale=0.8]{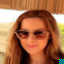}

}~~~~~~\subfloat[$5$$\times$$5$ red colour square trigger.\label{fig:red-solid}]{\includegraphics[scale=0.8]{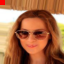}~~\includegraphics[scale=0.8]{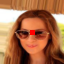}~~\includegraphics[scale=0.8]{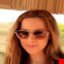}

}\caption{Different triggers (blue and green checkerboard and red trigger) with
different locations top-left, middle and bottom-right for CelebA dataset
are shown in Figure \ref{fig:cb-bg} and \ref{fig:red-solid} respectively.}
\vspace{-5bp}
\end{figure}
We conducted experiments using two types of triggers, checkerboard
of size $5$$\times$$5$ with blue and green colour and a red square
of size $5$$\times$$5$ on the CelebA dataset. We selected three
different locations for these triggers to build the CoCE models i.e.,
\emph{1. Top left with location as (0,0), 2. Middle with location
as (30,30), }and\emph{ 3. Bottom right with location as (59,59)} as
shown in Figures \ref{fig:cb-bg} and \ref{fig:red-solid}. The composite
concept we used is \emph{male eyeglass} and the settings of the experiments
are the same as the results reported in Table \ref{tab:auc-clip-sotas}.
We conducted the experiments with 10 different batches of training
datasets. We use 10 and 20 samples of positive and negative training
datasets.

\begin{table}[H]
\begin{centering}
\vspace{-4bp}
{\footnotesize{}}%
\begin{tabular}{ccccccc}
\toprule 
{\small{}Dataset} & Secondary & Trigger & Trigger & \multicolumn{3}{c}{Trigger location}\tabularnewline
 & concept & type & size & Left (0,0) & Middle (30,30) & Right (59,59)\tabularnewline
\midrule
\multirow{2}{*}{CelebA} & \multirow{2}{*}{Eyeglass} & Checker board & \multirow{2}{*}{$5$$\times$$5$} & 0.64$\pm$0.03 & \textbf{0.68$\pm$0.03} & 0.63$\pm$0.03\tabularnewline
\cmidrule{3-3} \cmidrule{5-7} 
 &  & Red square &  & 0.65$\pm$0.04 & \textbf{0.67$\pm$0.02} & 0.65$\pm$0.03\tabularnewline
\bottomrule
\end{tabular}{\footnotesize\par}
\par\end{centering}
\caption{Average AUC scores of CoCE models trained with checkerboard and red
color triggers of sizes $5$$\times$$5$ with different trigger locations
(top-left, middle, and bottom-right) on the image.\label{tab:trgr-1}}
\vspace{-1bp}
\end{table}
Table \ref{tab:trgr-1} reports the experiments when we use different
triggers with varying locations. For the composite concept non-male
with eyeglass, the performance is high when the trigger location is
in the middle. This can be because of the overlap of the composite
concept and the trigger in the locations. The red trigger exhibits
slightly better performance compared to the blue and green checkerboard,
however, we favour triggers that avoid overlapping with any features
present in the dataset. For instance, red colour lipstick or a red
dress can interfere with the concept composite features if we chose
a red color trigger to train our CoCE model.

\subsubsection{Few-shot analysis}

\begin{table}[H]
\begin{centering}
\vspace{-5bp}
{\footnotesize{}}%
\begin{tabular}{ccccccccc}
\toprule 
\multirow{2}{*}{{\scriptsize{}Dataset}} & {\scriptsize{}Secondary} & \multicolumn{7}{c}{{\scriptsize{}$\left[N_{p},N_{n}\right]$}}\tabularnewline
\cmidrule{3-9} 
 & {\scriptsize{}concept} & {\scriptsize{}{[}2, 4{]}} & {\scriptsize{}{[}5, 10{]}} & {\scriptsize{}{[}10, 20{]}} & {\scriptsize{}{[}20, 40{]}} & {\scriptsize{}{[}30, 60{]}} & {\scriptsize{}{[}40, 80{]}} & {\scriptsize{}{[}50, 100{]}}\tabularnewline
\midrule
\multirow{4}{*}{{\scriptsize{}CelebA}} & {\scriptsize{}Blond hair} & {\scriptsize{}0.71$\pm$0.02} & {\scriptsize{}0.73$\pm$0.02} & {\scriptsize{}0.73$\pm$0.03} & {\scriptsize{}0.73$\pm$0.02} & {\scriptsize{}0.73$\pm$0.03} & \textbf{\scriptsize{}0.79$\pm$0.04} & {\scriptsize{}0.75$\pm$0.04}\tabularnewline
\cmidrule{2-9} 
 & {\scriptsize{}Eyeglass} & {\scriptsize{}0.61$\pm$0.02} & {\scriptsize{}0.63$\pm$0.03} & {\scriptsize{}0.64$\pm$0.03} & {\scriptsize{}0.66$\pm$0.03} & {\scriptsize{}0.68$\pm$0.03} & {\scriptsize{}0.73$\pm$0.04} & \textbf{\scriptsize{}0.74$\pm$0.04}\tabularnewline
\cmidrule{2-9} 
 & {\scriptsize{}Paleskin} & {\scriptsize{}0.64$\pm$0.03} & {\scriptsize{}0.66$\pm$0.06} & {\scriptsize{}0.66$\pm$0.02} & {\scriptsize{}0.68$\pm$0.04} & {\scriptsize{}0.68$\pm$0.04} & {\scriptsize{}0.70$\pm$0.06} & \textbf{\scriptsize{}0.70$\pm$0.06}\tabularnewline
\cmidrule{2-9} 
 & {\scriptsize{}Wearing hat} & {\scriptsize{}0.75$\pm$0.03} & {\scriptsize{}0.75$\pm$0.01} & {\scriptsize{}0.76$\pm$0.02} & {\scriptsize{}0.79$\pm$0.02} & {\scriptsize{}0.82$\pm$0.03} & {\scriptsize{}0.81$\pm$0.03} & \textbf{\scriptsize{}0.82$\pm$0.04}\tabularnewline
\bottomrule
\end{tabular}{\footnotesize\par}
\par\end{centering}
\caption{Average AUC score of CoCE (10 runs) with varying number of positive
and negative training data. We used a checker board of size $5$$\times$$5$
with blue and green colour as our trigger for the CoCE models. \label{tab:ablation-few-shot-AUC}}
\vspace{-1bp}
\end{table}
We chose the composite concepts of CelebA dataset such as \emph{male
with blond hair}, \emph{male with eyeglass}, \emph{non-male with pale
skin}, and \emph{non-male with hat} to conduct the few-shot analysis
experiments. The associated secondary concepts, \emph{blondhair},
\emph{eyeglass}, \emph{paleskin} and \emph{wearing hat} are presented
in the Table \ref{tab:ablation-few-shot-AUC} for clarity. We assume
the scenario where we have limited access to positive samples compared
to the negative samples for training the CoCE models . We run each
composite concept 10 times with varying numbers of positive and negative
training sets. The mean and standard deviation reported over 10 runs
are shown in Table \ref{tab:ablation-few-shot-AUC}. It is evident
from the Table \ref{tab:ablation-few-shot-AUC} that the AUC scores
will improve with more samples from the positive and negative training
datasets. The values of $\left[N_{p},N_{n}\right]$ in each column
show the number of positive and negative samples we have used for
CoCE models.

\section{Conclusion}

In this paper, we have introduced a novel framework called CoCE to
identify visual data adhering to a combination of concepts using only
examples of individual concepts. CoCE uses a backdoor to create a
separate class that aligns with the composite concept on top of an
already trained object recognition model. The learning also utilizes
contrastive learning to learn the composite class using only a few
samples of positive and negative datasets, each corresponding to individual
concepts. Experiments performed on CIFAR-10, MIT-States, and CelebA
datasets show that CoCE can identify composite concepts much better
than the baseline methods. For future work, we will focus on developing
an optimized universal trigger for contrastive learning and enabling
CoCE for extracting more than one secondary concepts together.

\bibliographystyle{unsrtnat}
\phantomsection\addcontentsline{toc}{section}{\refname}\bibliography{references_1}

\begin{thebibliography}{36}
\providecommand{\natexlab}[1]{#1}
\providecommand{\url}[1]{\texttt{#1}}
\expandafter\ifx\csname urlstyle\endcsname\relax
  \providecommand{\doi}[1]{doi: #1}\else
  \providecommand{\doi}{doi: \begingroup \urlstyle{rm}\Url}\fi

\bibitem[Han et~al.(2019)Han, Mao, Gan, Tenenbaum, and Wu]{Han_etal_19Visual}
Chi Han, Jiayuan Mao, Chuang Gan, Josh Tenenbaum, and Jiajun Wu.
\newblock Visual concept-metaconcept learning.
\newblock \emph{Advances in Neural Information Processing Systems}, 32, 2019.

\bibitem[Hu et~al.(2022)Hu, Salcic, Dobbie, Chen, Sun, and Zhang]{Hu_etal_2022Membership}
Hongsheng Hu, Zoran Salcic, Gillian Dobbie, Jinjun Chen, Lichao Sun, and Xuyun Zhang.
\newblock Membership inference via backdooring.
\newblock \emph{arXiv preprint arXiv:2206.04823}, 2022.

\bibitem[Adi et~al.(2018)Adi, Baum, Cisse, Pinkas, and Keshet]{Adi_etal_18Turning}
Yossi Adi, Carsten Baum, Moustapha Cisse, Benny Pinkas, and Joseph Keshet.
\newblock Turning your weakness into a strength: Watermarking deep neural networks by backdooring.
\newblock In \emph{27th USENIX Security Symposium (USENIX Security 18)}, pages 1615--1631, 2018.

\bibitem[Li et~al.(2023{\natexlab{a}})Li, Zhu, Yang, Jiang, Wei, and Xia]{Li_etal_2023Black}
Yiming Li, Mingyan Zhu, Xue Yang, Yong Jiang, Tao Wei, and Shu-Tao Xia.
\newblock Black-box dataset ownership verification via backdoor watermarking.
\newblock \emph{IEEE Transactions on Information Forensics and Security}, 2023{\natexlab{a}}.

\bibitem[Shan et~al.(2020)Shan, Wenger, Wang, Li, Zheng, and Zhao]{Shan_etal_20Gotta}
Shawn Shan, Emily Wenger, Bolun Wang, Bo~Li, Haitao Zheng, and Ben~Y Zhao.
\newblock Gotta catch'em all: Using honeypots to catch adversarial attacks on neural networks.
\newblock In \emph{Proceedings of the 2020 ACM SIGSAC Conference on Computer and Communications Security}, pages 67--83, 2020.

\bibitem[Malinowski et~al.(2015)Malinowski, Rohrbach, and Fritz]{Malinowski_etal2015Ask}
Mateusz Malinowski, Marcus Rohrbach, and Mario Fritz.
\newblock Ask your neurons: A neural-based approach to answering questions about images.
\newblock In \emph{Proceedings of the IEEE international conference on computer vision}, pages 1--9, 2015.

\bibitem[Mei et~al.(2022)Mei, Mao, Wang, Gan, and Tenenbaum]{Mei_etal2022Falcon}
Lingjie Mei, Jiayuan Mao, Ziqi Wang, Chuang Gan, and Joshua~B Tenenbaum.
\newblock Falcon: fast visual concept learning by integrating images, linguistic descriptions, and conceptual relations.
\newblock \emph{arXiv preprint arXiv:2203.16639}, 2022.

\bibitem[Misra et~al.(2017)Misra, Gupta, and Hebert]{misra2017red}
Ishan Misra, Abhinav Gupta, and Martial Hebert.
\newblock From red wine to red tomato: Composition with context.
\newblock In \emph{Proceedings of the IEEE Conference on Computer Vision and Pattern Recognition}, pages 1792--1801, 2017.

\bibitem[Mancini et~al.(2021)Mancini, Naeem, Xian, and Akata]{mancini2021open}
Massimiliano Mancini, Muhammad~Ferjad Naeem, Yongqin Xian, and Zeynep Akata.
\newblock Open world compositional zero-shot learning.
\newblock In \emph{Proceedings of the IEEE/CVF conference on computer vision and pattern recognition}, pages 5222--5230, 2021.

\bibitem[Li et~al.(2022{\natexlab{a}})Li, Yang, Wei, Deng, and Yang]{li2022siamese}
Xiangyu Li, Xu~Yang, Kun Wei, Cheng Deng, and Muli Yang.
\newblock Siamese contrastive embedding network for compositional zero-shot learning.
\newblock In \emph{Proceedings of the IEEE/CVF conference on computer vision and pattern recognition}, pages 9326--9335, 2022{\natexlab{a}}.

\bibitem[Radford et~al.(2021)Radford, Kim, Hallacy, Ramesh, Goh, Agarwal, Sastry, Askell, Mishkin, Clark, et~al.]{Radford_etal_2021Learning}
Alec Radford, Jong~Wook Kim, Chris Hallacy, Aditya Ramesh, Gabriel Goh, Sandhini Agarwal, Girish Sastry, Amanda Askell, Pamela Mishkin, Jack Clark, et~al.
\newblock Learning {T}ransferable {V}isual {M}odels {F}rom {N}atural {L}anguage {S}upervision.
\newblock In \emph{International {C}onference on {M}achine {L}earning}, 2021.

\bibitem[Yang et~al.(2018)Yang, Lu, Lee, Batra, and Parikh]{Yang_etal_2018Graph}
Jianwei Yang, Jiasen Lu, Stefan Lee, Dhruv Batra, and Devi Parikh.
\newblock Graph {R-CNN} for {S}cene {G}raph {G}eneration.
\newblock In \emph{Proceedings of the European {C}onference on {C}omputer {V}ision}, 2018.

\bibitem[Gu et~al.(2017)Gu, Dolan-Gavitt, and Garg]{Gu_etal_17Badnets}
Tianyu Gu, Brendan Dolan-Gavitt, and Siddharth Garg.
\newblock Badnets: Identifying {V}ulnerabilities in the {M}achine {L}earning {M}odel {S}upply {C}hain.
\newblock \emph{arXiv preprint arXiv:1708.06733}, 2017.

\bibitem[Jha et~al.(2023)Jha, Hayase, and Oh]{jha2023label}
Rishi Jha, Jonathan Hayase, and Sewoong Oh.
\newblock Label poisoning is all you need.
\newblock \emph{Advances in Neural Information Processing Systems}, 36:\penalty0 71029--71052, 2023.

\bibitem[Saha et~al.(2020)Saha, Subramanya, and Pirsiavash]{Saha_etal_20Hidden}
Aniruddha Saha, Akshayvarun Subramanya, and Hamed Pirsiavash.
\newblock Hidden trigger backdoor attacks.
\newblock In \emph{Proceedings of the AAAI Conference on Artificial Intelligence}, volume~34, pages 11957--11965, 2020.

\bibitem[Chen et~al.(2017)Chen, Liu, Li, Lu, and Song]{chen2017targeted}
Xinyun Chen, Chang Liu, Bo~Li, Kimberly Lu, and Dawn Song.
\newblock Targeted {B}ackdoor {A}ttacks on {D}eep {L}earning {S}ystems using {D}ata {P}oisoning.
\newblock \emph{arXiv preprint arXiv:1712.05526}, 2017.

\bibitem[Doan et~al.(2021)Doan, Lao, Zhao, and Li]{Doan_etal_21Lira}
Khoa Doan, Yingjie Lao, Weijie Zhao, and Ping Li.
\newblock Lira: {L}earnable, {I}mperceptible and {R}obust {B}ackdoor {A}ttacks.
\newblock In \emph{Proceedings of the IEEE/CVF international conference on computer vision}, pages 11966--11976, 2021.

\bibitem[Nguyen and Tran(2020)]{nguyen2020input}
Tuan~Anh Nguyen and Anh Tran.
\newblock Input-{A}ware {D}ynamic {B}ackdoor {A}ttack.
\newblock \emph{Advances in Neural Information Processing Systems}, 33:\penalty0 3454--3464, 2020.

\bibitem[Gu et~al.(2019)Gu, Liu, Dolan-Gavitt, and Garg]{gu2019badnets}
Tianyu Gu, Kang Liu, Brendan Dolan-Gavitt, and Siddharth Garg.
\newblock Badnets: Evaluating {B}ackdooring {A}ttacks on {D}eep {N}eural {N}etworks.
\newblock \emph{IEEE Access}, 7:\penalty0 47230--47244, 2019.

\bibitem[Harikumar et~al.(2022)Harikumar, Rana, Do, Gupta, Zong, Susilo, and Venkastesh]{Harikumar_etal_22Defense}
Haripriya Harikumar, Santu Rana, Kien Do, Sunil Gupta, Wei Zong, Willy Susilo, and Svetha Venkastesh.
\newblock Defense against multi-target trojan attacks.
\newblock \emph{arXiv preprint arXiv:2207.03895}, 2022.

\bibitem[Wenger et~al.(2021)Wenger, Passananti, Bhagoji, Yao, Zheng, and Zhao]{wenger2021backdoor}
Emily Wenger, Josephine Passananti, Arjun~Nitin Bhagoji, Yuanshun Yao, Haitao Zheng, and Ben~Y Zhao.
\newblock Backdoor attacks against deep learning systems in the physical world.
\newblock In \emph{Proceedings of the IEEE/CVF Conference on Computer Vision and Pattern Recognition}, pages 6206--6215, 2021.

\bibitem[Harikumar et~al.(2021{\natexlab{a}})Harikumar, Do, Rana, Gupta, and Venkatesh]{harikumar2021semantic}
Haripriya Harikumar, Kien Do, Santu Rana, Sunil Gupta, and Svetha Venkatesh.
\newblock Semantic host-free trojan attack.
\newblock \emph{arXiv preprint arXiv:2110.13414}, 2021{\natexlab{a}}.

\bibitem[Wang et~al.(2019)Wang, Yao, Shan, Li, Viswanath, Zheng, and Zhao]{Wang_etal_19Neural}
Bolun Wang, Yuanshun Yao, Shawn Shan, Huiying Li, Bimal Viswanath, Haitao Zheng, and Ben~Y Zhao.
\newblock Neural {C}leanse: {I}dentifying and {M}itigating {B}ackdoor {A}ttacks in {N}eural {N}etworks.
\newblock In \emph{IEEE Symposium on Security and Privacy}, pages 707--723. IEEE, 2019.

\bibitem[Harikumar et~al.(2021{\natexlab{b}})Harikumar, Le, Rana, Bhattacharya, Gupta, and Venkatesh]{harikumar2021scalable}
H~Harikumar, Vuong Le, Santu Rana, S~Bhattacharya, Sunil Gupta, and Svetha Venkatesh.
\newblock Scalable {B}ackdoor {D}etection in {N}eural {N}etworks.
\newblock In \emph{Joint European Conference on Machine Learning and Knowledge Discovery in Databases}, pages 289--304. Springer, 2021{\natexlab{b}}.

\bibitem[Liu et~al.(2019)Liu, Lee, Tao, Ma, Aafer, and Zhang]{Liu_etal_19Abs}
Yingqi Liu, Wen-Chuan Lee, Guanhong Tao, Shiqing Ma, Yousra Aafer, and Xiangyu Zhang.
\newblock A{BS}: Scanning {N}eural {N}etworks for {B}ack-doors by {A}rtificial {B}rain {S}timulation.
\newblock In \emph{Proceedings of the ACM SIGSAC Conference on Computer and Communications Security}, pages 1265--1282, 2019.

\bibitem[Fu et~al.(2023)Fu, Zhang, Ji, Wang, Lin, Feng, and Yin]{fu2023freeeagle}
Chong Fu, Xuhong Zhang, Shouling Ji, Ting Wang, Peng Lin, Yanghe Feng, and Jianwei Yin.
\newblock $\{$FreeEagle$\}$: Detecting complex neural trojans in $\{$Data-Free$\}$ cases.
\newblock In \emph{32nd USENIX Security Symposium (USENIX Security 23)}, pages 6399--6416, 2023.

\bibitem[Zheng et~al.(2022)Zheng, Tang, Li, and Liu]{zheng2022pre}
Runkai Zheng, Rongjun Tang, Jianze Li, and Li~Liu.
\newblock Pre-activation distributions expose backdoor neurons.
\newblock \emph{Advances in Neural Information Processing Systems}, 35:\penalty0 18667--18680, 2022.

\bibitem[Li et~al.(2021)Li, Lyu, Koren, Lyu, Li, and Ma]{li_21neural}
Yige Li, Xixiang Lyu, Nodens Koren, Lingjuan Lyu, Bo~Li, and Xingjun Ma.
\newblock Neural {A}ttention {D}istillation: Erasing {B}ackdoor {T}riggers from {D}eep {N}eural {N}etworks.
\newblock In \emph{International Conference on Learning Representations}, 2021.

\bibitem[Garipov et~al.(2018)Garipov, Izmailov, Podoprikhin, Vetrov, and Wilson]{Garipov_etal_18Loss}
Timur Garipov, Pavel Izmailov, Dmitrii Podoprikhin, Dmitry~P Vetrov, and Andrew~G Wilson.
\newblock Loss surfaces, mode connectivity, and fast ensembling of dnns.
\newblock \emph{Advances in neural information processing systems}, 31, 2018.

\bibitem[Wu and Wang(2021)]{wu2021adversarial}
Dongxian Wu and Yisen Wang.
\newblock Adversarial neuron pruning purifies backdoored deep models.
\newblock \emph{Advances in Neural Information Processing Systems}, 34:\penalty0 16913--16925, 2021.

\bibitem[Li et~al.(2023{\natexlab{b}})Li, Lyu, Ma, Koren, Lyu, Li, and Jiang]{li2023reconstructive}
Yige Li, Xixiang Lyu, Xingjun Ma, Nodens Koren, Lingjuan Lyu, Bo~Li, and Yu-Gang Jiang.
\newblock Reconstructive neuron pruning for backdoor defense.
\newblock In \emph{International Conference on Machine Learning}, pages 19837--19854. PMLR, 2023{\natexlab{b}}.

\bibitem[Do et~al.(2022)Do, Harikumar, Le, Nguyen, Tran, Rana, Nguyen, Susilo, and Venkatesh]{Do_etal_2022Towards}
Kien Do, Haripriya Harikumar, Hung Le, Dung Nguyen, Truyen Tran, Santu Rana, Dang Nguyen, Willy Susilo, and Svetha Venkatesh.
\newblock Towards {E}ffective and {R}obust {N}eural {T}rojan {D}efenses via {I}nput {F}iltering.
\newblock In \emph{European Conference on Computer Vision}, pages 283--300. Springer, 2022.

\bibitem[Doan et~al.(2020)Doan, Abbasnejad, and Ranasinghe]{Doan_etal_20Februus}
Bao~Gia Doan, Ehsan Abbasnejad, and Damith~C Ranasinghe.
\newblock Februus: Input purification defense against trojan attacks on deep neural network systems.
\newblock In \emph{Annual computer security applications conference}, pages 897--912, 2020.

\bibitem[Gao et~al.(2019)Gao, Xu, Wang, Chen, Ranasinghe, and Nepal]{Gao_etal_19Strip}
Yansong Gao, Change Xu, Derui Wang, Shiping Chen, Damith~C Ranasinghe, and Surya Nepal.
\newblock {S}trip: {A} {D}efence {A}gainst {T}rojan {A}ttacks on {D}eep {N}eural {N}etworks.
\newblock In \emph{Proceedings of the 35th Annual Computer Security Applications Conference}, pages 113--125, 2019.

\bibitem[Li et~al.(2022{\natexlab{b}})Li, Bai, Jiang, Yang, Xia, and Li]{Li_etal_22Untargeted}
Yiming Li, Yang Bai, Yong Jiang, Yong Yang, Shu-Tao Xia, and Bo~Li.
\newblock Untargeted backdoor watermark: Towards harmless and stealthy dataset copyright protection.
\newblock \emph{Advances in Neural Information Processing Systems}, 35:\penalty0 13238--13250, 2022{\natexlab{b}}.

\bibitem[Sommer et~al.(2020)Sommer, Song, Wagh, and Mittal]{Sommer_etal_2020Towards}
David~Marco Sommer, Liwei Song, Sameer Wagh, and Prateek Mittal.
\newblock Towards probabilistic verification of machine unlearning.
\newblock \emph{arXiv preprint arXiv:2003.04247}, 2020.

\end{thebibliography}

\appendix
\newpage
\end{document}